\documentclass[letterpaper, 10 pt, conference]{ieeeconf}  

\IEEEoverridecommandlockouts                              

\usepackage{graphics} 
\usepackage{amsmath} 
\usepackage{booktabs}
\usepackage{multirow}
\usepackage{graphicx}
\usepackage{fancyhdr}
\usepackage{hyperref}

\begin{document}
\title{\LARGE \bf
SeqNet: Learning Descriptors for Sequence-based Hierarchical Place Recognition
}

\author{Sourav Garg and Michael Milford
\thanks{The authors are with the QUT Centre for Robotics and School of Electrical Engineering and Robotics, QUT, Brisbane, Australia.
The authors acknowledge continued support from the Queensland University of Technology (QUT) through the Centre for Robotics.
        \tt\small s.garg@qut.edu.au}}%

\maketitle
\thispagestyle{fancy}
\pagestyle{plain}

\begin{abstract}

Visual Place Recognition (VPR) is the task of matching current visual imagery from a camera to images stored in a reference map of the environment. While initial VPR systems used simple direct image methods or hand-crafted visual features, recent work has focused on learning more powerful visual features and further improving performance through either some form of sequential matcher / filter or a hierarchical matching process. In both cases the performance of the initial single-image based system is still far from perfect, putting significant pressure on the sequence matching or (in the case of hierarchical systems) pose refinement stages. In this paper we present a novel hybrid system that creates a high performance initial match hypothesis generator using short learnt sequential descriptors, which enable selective control sequential score aggregation using single image learnt descriptors. Sequential descriptors are generated using a temporal convolutional network dubbed SeqNet, encoding short image sequences using 1-D convolutions, which are then matched against the corresponding temporal descriptors from the reference dataset to provide an ordered list of place match hypotheses. We then perform selective sequential score aggregation using shortlisted single image learnt descriptors from a separate pipeline to produce an overall place match hypothesis. Comprehensive experiments on challenging benchmark datasets demonstrate the proposed method outperforming recent state-of-the-art methods using the same amount of sequential information. Source code and supplementary material can be found at \url{https://github.com/oravus/seqNet}.

\end{abstract}

\section{INTRODUCTION}

Visual Place Recognition (VPR) under extreme appearance variations is a challenging task. Researchers have explored a variety of methods to deal with this problem ranging from traditional hand-crafted techniques~\cite{cummins2008fab,milford2012seqslam} to modern deep learning-based solutions~\cite{arandjelovic2016netvlad,garg19Semantic, sarlin2018bleveraging}. Many of these systems aim to push the performance of single image based place recognition by learning better image representations~\cite{arandjelovic2016netvlad, chen2017deep} and matchers~\cite{sarlin2020superglue}. To further improve the accuracy of such techniques, researchers have also explored the use of sequential information inherent within the problem of mobile robot localization. 

One of the most common uses of sequential information in VPR is to leverage the \textit{order} in which the visual information is accrued when a robot revisits a place. A well-known approach to this is based on constructing a `similarity matrix'~\cite{ho2007detecting} between the reference map and query observations where each entry of the matrix is typically computed through similarity between single image descriptors of images. The matching sub-sequences in the matrix can then be searched by aggregating the sequences of similarity scores particularly along the diagonal~\cite{milford2012seqslam, naseer2014robust, lynen2014placeless}. However, such sequence matching techniques have drawbacks: a) sequence score aggregation cannot get rid of high-confidence false matches of underlying single image descriptors without accessing sufficient additional sequential information and b) sequence searching within the whole database can be inefficient, typically scaling linearly with both the size of the reference map and the length of the image sequence.

\begin{figure}
    \centering
    \includegraphics[width=0.45\textwidth]{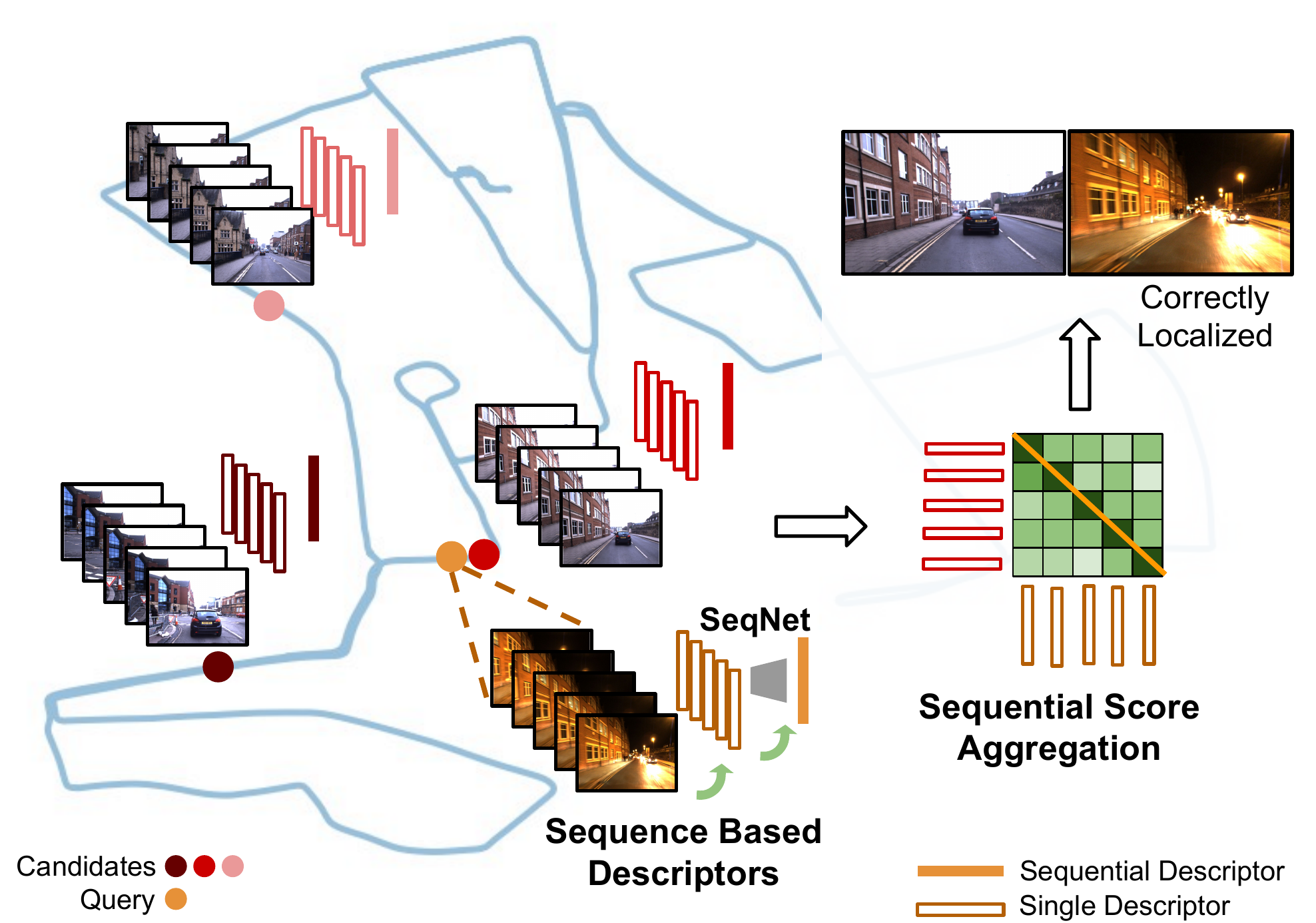}
    \caption{Sequence-based hierarchical visual place recognition. We propose SeqNet to learn short sequential descriptors that generate high performance initial match candidates and enable selective control sequence score aggregation using single image learnt descriptors.}
    \label{fig:front-page}
\end{figure}

\begin{figure*}[t]
    \centering
    \includegraphics[width=0.85\textwidth]{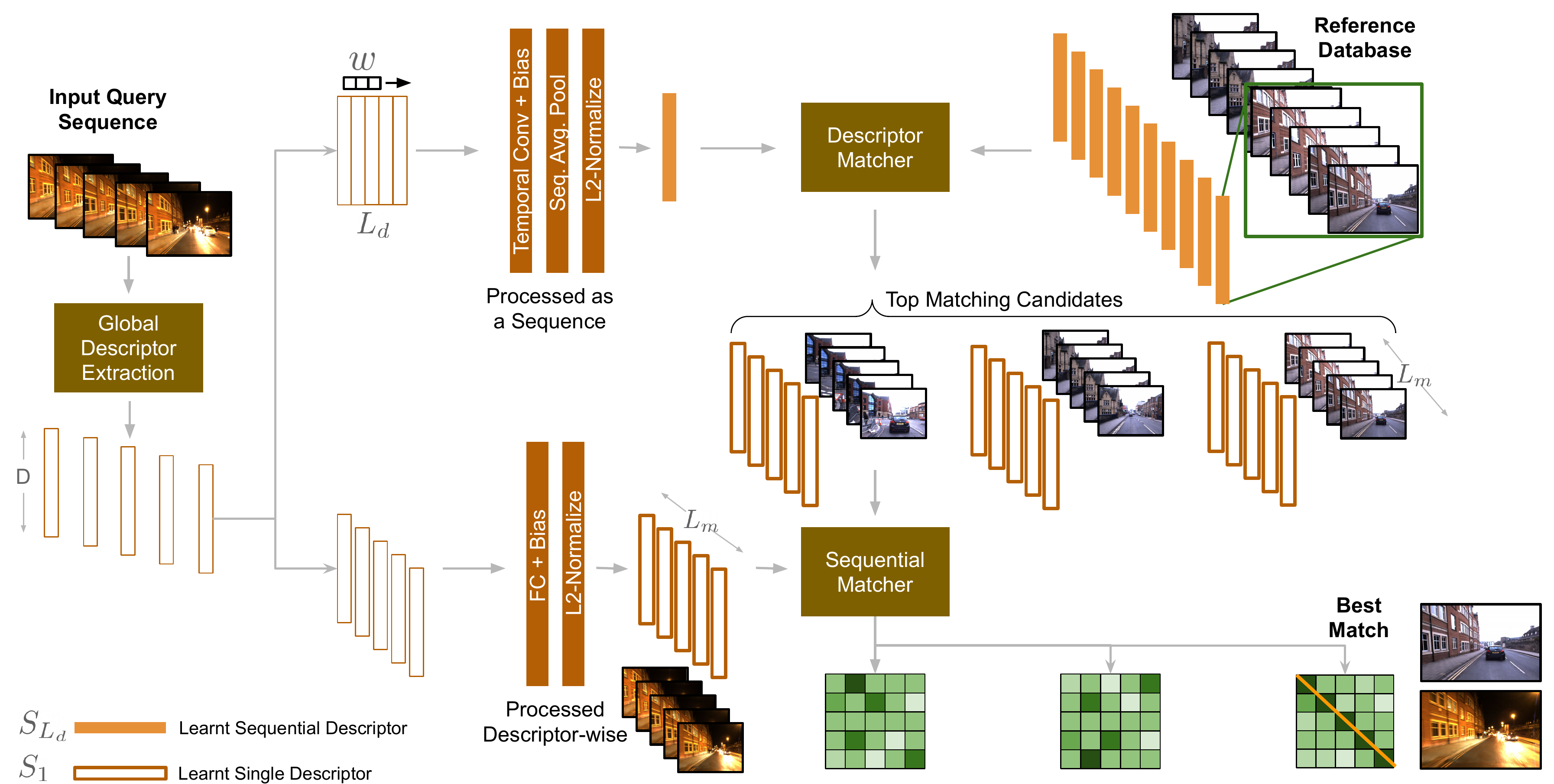}
    \caption{Our proposed hierarchical VPR pipeline using SeqNet to generate top match hypotheses through a sequential descriptor, followed by sequence matching of learnt single image descriptors.}
    \label{fig:schematic}
\end{figure*}

In this paper, we explore the use of temporal information through \textit{short} sequences of images whilst addressing the limitations of existing sequence-based VPR technique. For a fixed sequential span, we propose a hierarchical place recognition solution (see Figure~\ref{fig:front-page}) based on two complementary modules: i) a \textit{learnt sequential descriptor} capable of generating highly-accurate top match hypotheses and ii) a \textit{learnt single image descriptor} used in conjunction with \textit{sequential score aggregation} to precisely re-rank the top candidates by enforcing sequential order. Our proposed sequential descriptor-based shortlisting not only reduces the computational burden of sequence matching by a significant margin, it also eliminates the false positives to which single image descriptors have limited robustness. This becomes evident from our experimental results where we demonstrate that for a fixed sequential span, our proposed pipeline achieves similar or superior performance which is neither consistently achievable by sequential descriptors alone nor by sequential matching on top of vanilla or learnt single image descriptors, despite searching through the whole reference database.

We make the following contributions:
\begin{itemize}
    \item a novel spatial representation, SeqNet, based on sequential imagery, learnt using a single temporal convolutional layer and triplet loss;
    \item a low-latency state-of-the-art Hierarchical Visual Place Recognition (HVPR) system that combines the robustness of a sequential descriptor with the order-enforcing nature of a sequential matcher;
    \item detailed ablations and insights into the properties of our proposed system including recognizing places from a database traversed in reverse direction; and
    \item publicly available source code and supplementary results and visualizations to foster further research in this direction\footnote{\url{https://github.com/oravus/seqNet}}.
\end{itemize}

The paper proceeds as follows. Section~\ref{RELATEDWORK} briefly describes relevant Visual Place Recognition (VPR) research, focusing on sequential filtering and description techniques. In Section~\ref{PROPOSEDAPPROACH} we describe the key single and sequential descriptor learning processes including presenting \textit{SeqNet} and its integration in a new Hierarchical Visual Place Recognition (HVPR) framework. Section~\ref{EXPERIMENTALSETUP} describes a set of experiments with challenging benchmark datasets, with results presented in Section~\ref{RESULTS}. Finally we conclude in Section~\ref{DISCUSSION} with discussion and identification of the most promising areas of future work driven by the insights generated here.

\setlength{\topmargin}{-24pt}
\setlength{\headheight}{0pt}
\section{RELATED WORK}
\label{RELATEDWORK}
\subsection{Visual Place Recognition}
Visual place recognition has been extensively studied in the past where researchers have focused on several aspects of this problem including large scale appearance-based methods like FAB-MAP~\cite{cummins2008fab}, robust local~\cite{neubert2015superpixel}, and global feature representations~\cite{arroyo2014fast, naseer2014robust}, efficient retrieval~\cite{liu2012indexing, jegou2010aggregating, garg2020fast}, and biologically-inspired techniques~\cite{zaffar2020memorable}.

In the last decade, deep learning-based techniques have come to the fore in pushing the boundaries of viewpoint- and appearance-robust VPR. Some of the notable methods include end-to-end global representation learning~\cite{arandjelovic2016netvlad, chen2017deep}, use of learnt semantics~\cite{garg2018lost,garg2020semantics}, night-to-day image translation~\cite{anoosheh2019night, porav2018adversarial}, teacher-student networks~\cite{sarlin2018bleveraging} and reinforcement learning for navigation~\cite{chancan2019visual}. However, the majority of these methods are designed for single images.

\subsection{Sequential Score Aggregation}
Visual place recognition techniques have been shown to benefit from the use of sequence-based matching~\cite{ho2007detecting}, especially under the challenges of extreme appearance variations~\cite{milford2012seqslam}. Follow-up research has focused on improving these sequence score aggregation methods with the use of odometry~\cite{pepperell2014all}, camera velocity-robust sequence searching~\cite{naseer2014robust, vysotska2015efficient}, hashing based match selection~\cite{vysotska2017relocalization}, handling different routes~\cite{vysotska2019effective}, using temporal information within a diffusion process~\cite{zhang2019graph} and trajectory-based attention learning for SLAM~\cite{parisotto2018global}. However, most of these methods operate on the matching scores obtained from the underlying \textit{single image} descriptors.

\subsection{Sequential Descriptors}
While representing single images has been extensively studied in the literature, the use of temporal or sequential information to form a single compact representation has received limited attention in robotics, although numerous spatio-temporal description techniques exist in related areas of research~\cite{jalal2017robust, girdhar2017actionvlad, wu20193, Dai2017a}. Many of these methods learn temporal dynamics using RNN, LSTM~\cite{hochreiter1997long}, GRU~\cite{chung2014empiricalC} and more recently Temporal Convolution Networks~\cite{bai2018empirical}. In the context of VPR, researchers have explored learning spatio-temporal landmarks~\cite{johns2011place}, coresets-based summarization~\cite{volkov2015coresets}, depth-based topometric representation~\cite{garg2019look}, bio-inspired discriminative learning of memory cells~\cite{nguyen2013spatio} and multiple descriptors based grouping~\cite{zhang2016robust}, and more recently 3D information based descriptors~\cite{angelina2018pointnetvlad, du2020dh3d, oertel2020augmenting}.

Recently, \cite{facil2019condition} presented sequential representations learnt end-to-end using three different techniques: grouping (concatenation), fusion and recurrent. Concatenation of descriptors within a sequence has also been considered in an earlier work~\cite{arroyo2015towards}, where binarization and FLANN-based matching was employed for efficient place recognition. Similarly, recurrence has been employed in~\cite{neubert2019neurologically} for learning a topological map of the environment through single image descriptors. In this work, we consider a similar approach of decoupling single image global descriptor extraction and learning from sequential information. Unlike concatenation and recurrence based techniques that implicitly enforce strict temporal order, we employ 1D temporal convolutions for learning sequential descriptors and then hierarchically combine it with an explicit order-enforcing sequence score aggregation technique. More recently, \cite{garg2020delta} proposed Delta Descriptors as a sequential representation which adapts the existing single image descriptors in an unsupervised manner. However, it requires relatively long sequences and order-preserved route traversals; in this paper, we consider much shorter sequences, aiming for high localization performance with low latency.

\section{PROPOSED APPROACH}

Here we describe the SeqNet including the single and sequential descriptor learning processes and their place in the overall Hierarchical Visual Place Recognition structure. 

\label{PROPOSEDAPPROACH}

\subsection{SeqNet}
We propose a temporal convolutional network, dubbed SeqNet, to learn a spatial representation of the environment using sequentially-observed places. Given a sequence of RGB images, we first obtain the corresponding sequence of single-image descriptors using a state-of-the-art global descriptor extractor, NetVLAD~\cite{arandjelovic2016netvlad}. This sequence of descriptors is then fed through SeqNet as shown in the top row of Figure~\ref{fig:schematic} to obtain a single compact representation, represented as solid orange descriptor, referred to as $S_{L_d}$ and explained in subsequent sections.

\subsubsection{Network Architecture}
SeqNet is composed of a Temporal Convolutional (TC) layer (with bias, without padding), a Sequence Average Pooling (SAP) layer and an L2-Normalize layer. For the TC layer, we use a 1-D filter of length $w$ that operates (with stride 1) on an input \textit{sequence} of descriptors, that is, a tensor of size ($L_d\times D$) where $D$ represents the number of descriptor dimensions and $L_d$ represents the sequence length\footnote{We use $d$ and $m$ as subscript for sequence length $L$ to distinguish between its context being that of a sequential $d$escriptor or a sequential $m$atcher.}. The 1-D temporal convolutions operate in the sequence dimension of the input tensor, as shown in the top row of Figure~\ref{fig:schematic}. The descriptor dimension of the input tensor forms the input channels (feature maps) for the TC layer, with number of output channels set to $D$ which allows for fair benchmarking against similar size descriptors. Thus, the convolutional kernel tensor is of size $D\times w \times D$. The output of TC layer is a sub-sequence which is converted into a \textit{single} descriptor via SAP layer which performs averaging along the sequence (temporal) dimension, analogous to the GAP layer in the image space. While TC enables interaction among input sequence of descriptors within a local temporal window (equivalent to $w$), SAP ensures an output of sequence length 1. Since cosine distance is typically used for high-dimensional descriptor comparisons~\cite{sunderhauf2015performance,neubert2019neurologically}, we use the final L2-Normalize layer and Euclidean distance to mimic the same behaviour~\cite{qian2004similarity}. This is similar to the inter-normalization step used in~\cite{arandjelovic2016netvlad} before computing descriptor distances for triplet loss.

\subsubsection{Triplet Loss}
We train SeqNet using a set of reference and query databases, where for each query considered as an anchor $X_a$, its positives $X_p$ and negatives $X_n$ are generated from the reference database using a pre-defined localization radius. Similar to the training regime of NetVLAD~\cite{arandjelovic2016netvlad}, we use max-margin triplet loss as described below:
\begin{equation}
    l = max(\lVert X_a-X_p\rVert_2 - \lVert X_a-X_n\rVert_2 + \alpha, 0)
\end{equation}
where $\alpha$ is the desired margin between the positives and negatives in the descriptor space.

\subsection{Hierarchical Visual Place Recognition (HVPR)}
Hierarchical approaches for visual place recognition have been explored in the past in a variety of contexts. Distinct from existing methods, we define hierarchy with regards to the use of temporal information such that a ``summary'' sequential descriptor is used to shortlist matching candidates for subsequent sequence score aggregation. To achieve this, we consider a slight variant of our SeqNet model to additionally learn (adapt) single image descriptors, where sequence length for SeqNet is set to 1.

SeqNet transforms a sequence of input image descriptors ($L_d\times D$) into a $D$-dimensional sequential descriptor, referred to as $S_{L_d}$. Additionally, we train SeqNet with $L_d$=1 and $w$=1 to obtain a $D$-dimensional \textit{learnt} single image descriptor $S_1$; this is equivalent to learning a linear transformation (matrix multiplication and addition, similar to PCA transformation) of the input single image descriptors using a fully-connected layer and bias, as shown in Figure~\ref{fig:schematic}.

With the help of $S_{L_d}$ and $S_1$, we propose a hierarchical visual place recognition system that combines the robustness of learnt single and sequential descriptors with order-enforcing sequential score aggregation. For a given sequence of query images, we use its $S_{L_d}$ descriptor and Euclidean distance based matching to retrieve top K nearest neighbors from the reference database. 

\begin{equation}
    p_{ij} =  \lVert S_{L_d}^i - S_{L_d}^{j}\rVert_2 \quad \forall \quad j \in N_{db}
\end{equation}
where $N_{db}$ is the size of reference database and $p_{ij}$ is the descriptor distance of query sequential descriptor at index $i$ from the reference sequential descriptor at index $j$. A list of top candidates $R_i$ based on K lowest distance values is then considered for re-ranking using a simplified version of SeqSLAM~\cite{milford2012seqslam}, operating directly on descriptor distances without its velocity searching. This is referred to as SeqMatch in this paper, see Equation~\ref{eq:seqMatch}. For each of the top K candidates, we consider a $L_m$-length sequence of learnt single-image descriptors ($S_1$) and compare them with the learnt single-image descriptors of the query sequence:

\begin{equation}
    q_{ik} = \sum_{t=0}^{L_m-1} \lVert S_{1}^{i-t} - S_{1}^{k-t}\rVert_2  \quad \forall \quad k \in R_i
    \label{eq:seqMatch}
\end{equation}
where $q_{ik}$ is the sequence score between the query sequence at index $i$ and the top matching reference candidate $k$. With the assumption of one-to-one correspondence between the reference and query sequence, the sequence score will be minimized for correctly ordered images. Thus, the minimum scoring candidate is selected as the final match.

\section{EXPERIMENTAL SETUP}
\label{EXPERIMENTALSETUP}
\subsection{Datasets}
We used various outdoor benchmark datasets from diverse environment types: urban city traverses where appearance conditions vary due to day-night cycles; a rail journey where appearance conditions vary due to seasonal cycles; and multiple city street sequences captured under a mixed variety of appearance conditions, predominantly from daytime.

\subsubsection{Urban City - Day vs Night}
We used sequential imagery and GPS data from two different cities: Oxford and Brisbane to validate the generalization of our proposed method across extreme day-night variations. From both the cities, we used one daytime traverse as the reference database and one nighttime traverse as the query database. For Oxford, we used the left stereo images from the traverse ids: 2015-03-17-11-08-44 and 2014-12-16-18-44-24 of the Oxford Robotcar dataset~\cite{maddern20171} where each is around 10 km long with 30K images. For Brisbane, we used the City Loop and City Loop Night traverses as described in~\cite{p1007michael2020} where each is around 38 km long with 25K images. We use models trained on one city to test performance on the other city.

\subsubsection{Rail Journey - Seasonal Variations}
The Nordland dataset~\cite{sunderhauf2013we} comprises $728$ km rail journey across vegetative open environment captured under four seasons. We use the Summer-Winter pair for training and validation, and use Spring-Fall pair for testing. We remove the image frames where the vehicle was stationary or passing through tunnels~\cite{sunderhauf2015performance}. 

\subsubsection{Mapillary Street Level Sequences (MSLS)} MSLS~\cite{warburg2020mapillary} is a recently released dataset for benchmarking sequence-based place recognition. It comprises image sequences from a diverse set of cities, captured under a variety of appearance conditions, attributed to variations in weather, season, structure and viewpoint. In our experiments, we used Melbourne for training and Amman, Boston, San Francisco and Copenhagen for testing.

\subsection{Data Pre-Processing}
The focus of this research is to explore how sequential information can be best exploited for VPR. For this purpose, we use pre-computed image descriptors as input to SeqNet. Although raw RGB images with global descriptor extractor can be used as a back-end for end-to-end training, we have not considered that setup in this study. We used NetVLAD~\cite{arandjelovic2016netvlad}\footnote{We provide additional results for our proposed methods using another descriptor type in the supplementary material.} as our global image descriptor (after PCA), processing down-sampled images of size $640\times 320$\footnote{For the Oxford Robotcar dataset, we additionally removed the car hood (bottom $160$ rows of pixels from original imagery) before resizing.} to obtain $D$-dimensional descriptors ($D = 4096$). Given a $L_d$-length sequence of images, a corresponding $L_d \times D$ size tensor is thus obtained from the global descriptor extractor and fed to SeqNet. For the urban city datasets, the traverses were pre-processed to keep an approximate 2m frame separation. Since the Brisbane City Loop dataset was captured at a lower frame rate, frame separation for this dataset was variable but kept to be at least 2m. For MSLS, sequences were used as such. Furthermore, for testing on Oxford dataset, we consider two sampling scenarios: Fixed Distance (FD) with 2m frame separation as used during training and Fixed Time (FT) where regardless of geographical separation every $6^{th}$ frame is considered to keep the split size similar to that of FD. The latter is used to test robustness against camera velocity variations. For all the datasets, we only consider images where a valid sequence (of length $5$) can be centered. We provide further details regarding data splits and training parameters in the supplementary material.

\subsection{Evaluation}
\subsubsection{Recall@K}
VPR techniques typically form a part of localization pipelines for generating location priors, for example, in visual SLAM~\cite{mur2017orb} and 6-DoF localization~\cite{sarlin2019coarseC}. Such localization pipelines require high recall performance for VPR as the following 3D pose estimation modules often have high precision in selecting the best match~\cite{sarlin2020superglue}. Hence, we use Recall@K as the performance metric. For a given localization radius, Recall@K is defined as the ratio of correctly retrieved queries within the top K predictions to the total number of queries. We use a localization radius of $10$ meters, $20$ meters and $1$ frame respectively for Oxford, Brisbane/MSLS and Nordland datasets.

\newcommand{\scaleOne}{0.35}
\begin{figure*}
    \centering
    \begin{tabular}{cccc}
        \multicolumn{4}{c}{\includegraphics[scale=0.35]{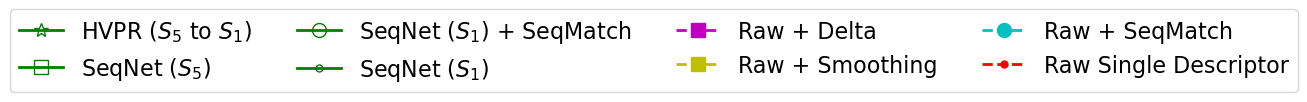}} \\
        \includegraphics[scale=\scaleOne]{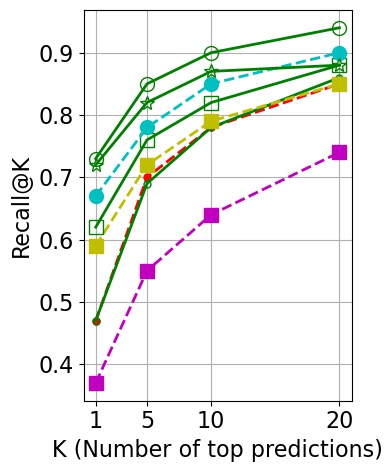} &
        \includegraphics[scale=\scaleOne]{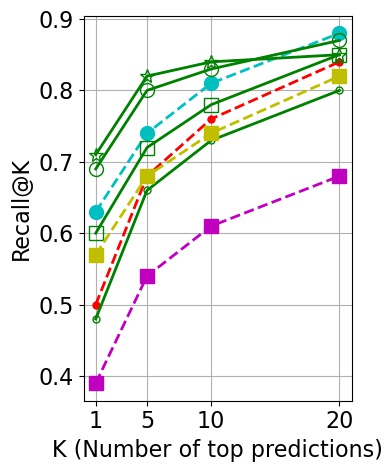} &
        \includegraphics[scale=\scaleOne]{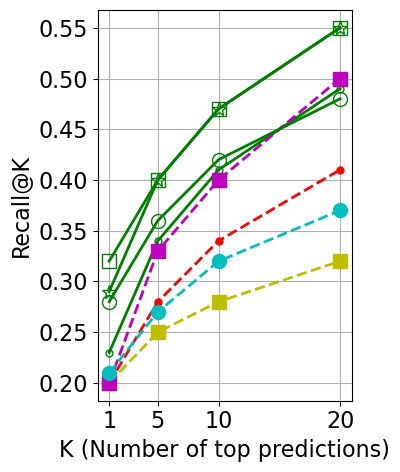} &
        \includegraphics[scale=\scaleOne]{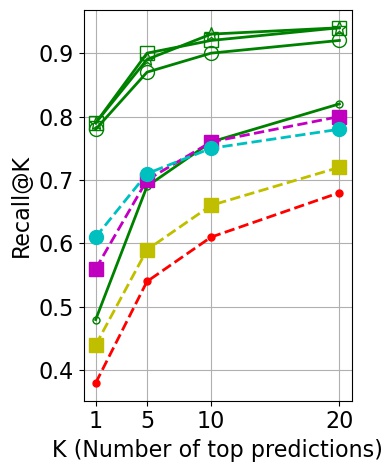} \\
        (a) Ox-FD (Day-Night) & (b) Ox-FT (Day-Night) & (c) Bne (Day-Night) & (d) Nord (Spr-Fal) \\ 
    \end{tabular}
    \caption{Recall performance on Oxford (Day-Night) with (a) Fixed-Distance (FD) sampling and (b) Fixed-Time (FT) sampling using Brisbane (Day-Night) trained model; (c) Brisbane (Day-Night) using Oxford (Day-Night) trained model; and (d) Nordland (Spring-Fall) with a model trained on Nordland (Summer-Winter).}
    \label{fig:mainResults1}
\end{figure*}

\begin{figure*}
    \centering
    \begin{tabular}{cccc}
       \includegraphics[scale=\scaleOne]{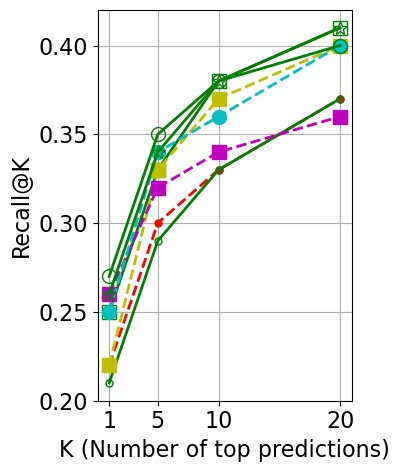} &
        \includegraphics[scale=\scaleOne]{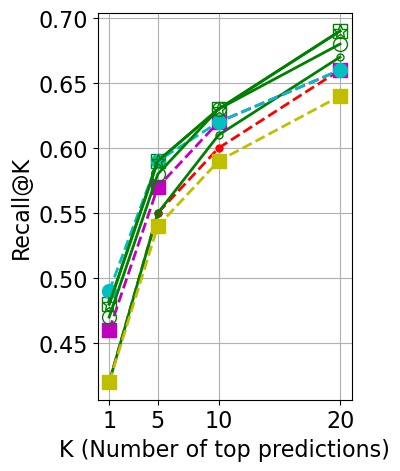} &
        \includegraphics[scale=\scaleOne]{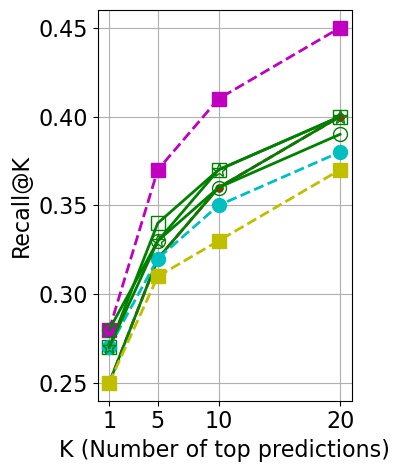} &
        \includegraphics[scale=\scaleOne]{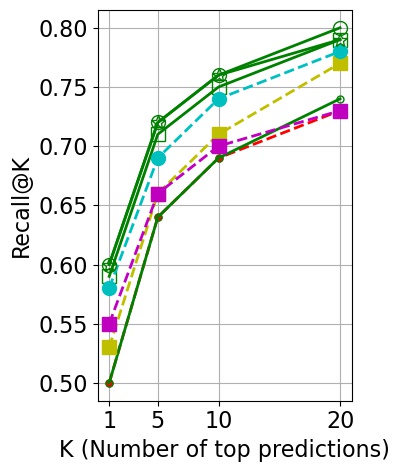} \\
        (a) Amman & (b) San Francisco & (c) Boston & (d) Copenhagen \\ 
    \end{tabular}
    \caption{Recall performance on MSLS: (a) Amman, (b) San Francisco, (c) Boston and (d) Copenhagen using a model trained on MSLS Melbourne.}
    \label{fig:mainResults2}
\end{figure*}

\subsubsection{Ablations and Baselines}
We first present ablations across different methods of using sequential information of place recognition. We compare against a) \textit{Single Image} descriptor (NetVLAD in this case) on top of which other methods are defined: b) \textit{Smoothing}, where a given descriptor is averaged within a temporal window (this is the same as the Sequence Average Pooling layer and equivalent to using SeqNet without temporal convolutions), c) \textit{Delta Descriptors}~\cite{garg2020delta}, where difference descriptors are computed on top of smoothed descriptors within a temporal window, d) \textit{SeqMatch}~\cite{garg2020fast}, where single image descriptors are compared between reference and query sequences to generate match scores which are then averaged to obtain the best match (see Eq.~\ref{eq:seqMatch}). All aforementioned descriptors are L2-normalized before any descriptor comparison.

Furthermore, we compare our proposed methods against single image descriptors: DenseVLAD~\cite{torii201524} and AP-GeM~\cite{revaud2019learning} in addition to NetVLAD~\cite{arandjelovic2016netvlad} and sequence-based place recognition methods: Graph-based Image Sequence Matcher~\cite{vysotska2015efficient}, referred to as GISM, and Graph-based Relocalization using Hashing~\cite{vysotska2017relocalization}, referred to as GRH with two types of hashing - LSH and DH, and the combination of Delta Descriptors with SeqMatch as proposed in~\cite{garg2020delta}. Note that GISM, GRH:LSH and GRH:DH operate in a `continuous' manner, that is, no sequence length parameter is needed. Considering the task of global re-localization, we constrain their sequence matching method to querying sequences of length $5$ independently so that all comparison methods use the same amount of sequential information. In the supplementary material, we provide additional results and comparisons using longer sequences.

\section{RESULTS}
\label{RESULTS}
\subsection{Urban City - Day vs Night}
Figure~\ref{fig:mainResults1} (a)-(c) show performance trends for VPR across day-night cycles for the city datasets where a SeqNet model trained on one city is tested on the other city. It can be observed that in all three cases the proposed sequential descriptor (green hollow squares) achieves superior performance as compared to other sequential descriptors: Smoothing and Delta. Further, HVPR (green stars) performs close to $S_1$+SeqMatch (green hollow circles) in (a) and surpasses it when variable camera velocity is considered in (b) and (c). Note that this performance gain is also accompanied by computational gain as only top 20 matches from $S_5$ are considered for sequence score aggregation.

In general, sequence score aggregation when operating on the whole database requires additional compute and cannot rectify high-confidence perceptual aliasing induced by the underlying descriptor unless a longer sequence is used. Our proposed HVPR pipeline fixes both these issues: firstly, the number of sequence matching operations are significantly reduced and secondly, the top-K candidates generated by SeqNet ($S_5$) are more accurate than those generated by SeqNet ($S_1$), thus eliminating those highly-confident false matches induced by $S_1$ which sequence matching by itself could not have fixed. 

\subsection{Rail Journey - Seasonal Variations}
Figure~\ref{fig:mainResults1} (d) shows performance trends for the Nordland dataset where our SeqNet model trained with Summer-Winter conditions is tested on Spring-Fall conditions. It can be observed that the training generalizes well across different scene appearance and geographical conditions. Furthermore, our proposed HVPR pipeline and sequential descriptor ($S_5$) achieve superior recall performance for all K values, outperforming both SeqMatch combinations (green and cyan circles).

\subsection{Mapillary Street Level Sequences (MSLS)}
Figure~\ref{fig:mainResults2} shows recall performance on four cities of the MSLS dataset using a model trained on one city (Melbourne). Unlike Day-Night and cross-season benchmarking in previous subsections, MSLS offers much more variability in terms of weather, structure, viewpoint and domain changes~\cite{warburg2020mapillary}. This is reflected by relatively thin performance margins between different methods in Figure~\ref{fig:mainResults2}. It can be observed that the use of sequential information with $S_5$, $S_1$+SeqMatch and HVPR leads to roughly similar performance but consistently better than other methods, except on the Boston dataset where Delta performs exceptionally well. The overall results on MSLS highlight the benefits of SeqNet and HVPR as compared to the use of sequential information through other means. The absolute performance bottleneck possibly lies in the challenging conditions of the dataset, where both a better underlying single image descriptor (e.g. NetVLAD trained on MSLS~\cite{warburg2020mapillary} or DenseVLAD~\cite{torii201524}) or better sequential description and matching architectures could potentially improve performance.

\subsection{Training Generalization}
We did not observe any one single trained model to perform well on \textit{all} the datasets, which can be attributed to lack of training on a possible combination set of different environment categories (city/man-made vs rail/natural) and appearance conditions (day-night and summer-winter). 
However, our results on Oxford and Brisbane demonstrate generality across different cities for a similar set of appearance conditions, that is day vs night. Similarly, results presented on Nordland demonstrate generality across different appearance conditions pairs, that is, spring-fall vs winter-summer. Finally, evaluation on MSLS shows that a model trained on one Australian city generalizes well across different cities from around the world.

\subsection{Comparison to other methods}
Table~\ref{tab:perfComp} shows performance comparisons on Oxford-FD and Brisbane dataset (same train/test settings as in Figure~\ref{fig:mainResults1}) for different method types including single image descriptors, sequential descriptors, sequence score aggregation methods and hierarchical methods.  In single image methods, DenseVLAD outperforms others, meaning SeqNet models could be trained on this method for further performance improvement. In sequential descriptors, SeqNet ($S_5$) achieves superior performance as compared to Smoothing and Delta. For sequence score aggregation methods, $S_1$+SeqMatch achieves superior performance. Unlike SeqMatch, GISM is capable of matching sequences with variable camera velocity, thus leading to relatively high performance on Brisbane dataset than the Oxford dataset where SeqMatch is able to exploit the same velocity constraint. The low performance numbers for GRH can be attributed to hashing which trades off performance for computation time. Finally, for hierarchical methods, obtaining shortlist candidates from Delta or NetVLAD does not help attain superior performance as achieved by HVPR ($S_5$ to $S_1$).

\begin{table}
\caption{Performance Comparison - Oxford \& Brisbane (Day vs Night): Recall@K}
    \centering
    \begin{tabular}{l ll}
        \toprule
        {} & \multicolumn{2}{c}{\textbf{Recall @ 1/5/20}} \\
        \cmidrule(lr{0.75em}){2-3}
        \textbf{Method}  & Oxford-FD & Brisbane \\
        \midrule

\textbf{Single Image Descriptors:} & &\\
        AP-GeM~\cite{revaud2019learning}  & 
 0.36/0.59/0.78 &
 0.21/0.33/0.48\\
 
        DenseVLAD~\cite{torii201524}  & 
 \textbf{0.50}/0.65/0.77 &
 \textbf{0.33}/\textbf{0.43}/\textbf{0.54}\\
 
        NetVLAD~\cite{arandjelovic2016netvlad}  & 
 0.47/\textbf{0.70}/0.85 &
 0.20/0.28/0.41\\

        SeqNet ($S_1$)  & 
        0.47/0.69/\textbf{0.86} &
        0.23/0.34/0.49\\

\midrule      
\textbf{Sequential Descriptors:} & & \\
        Smoothing~\cite{garg2020delta}   & 
        0.59/0.72/0.85 & 
        0.20/0.25/0.32\\

        Delta~\cite{garg2020delta}  & 
        0.37/0.55/0.74 &
        0.20/0.33/0.50\\
        
        SeqNet ($S_5$)  &  
        \textbf{0.62}/\textbf{0.76}/\textbf{0.88} &
        \textbf{0.32}/\textbf{0.40}/\textbf{0.55}\\
\midrule  
\textbf{Sequential Score Aggregation:} & & \\
        SeqMatch~\cite{milford2012seqslam}   & 
        0.67/0.78/0.9 & 
        0.21/0.31/0.37\\
        
        Delta + SeqMatch~\cite{garg2020delta}  & 
        0.64/0.81/0.91 &
        0.23/0.33/\textbf{0.48}\\

         SeqNet ($S_1$) + SeqMatch~\cite{milford2012seqslam}   & 
      \textbf{0.73}/\textbf{0.85}/\textbf{0.94} &
      \textbf{0.28}/\textbf{0.36}/\textbf{0.48}\\
      
         SeqNet ($S_1$) + GISM~\cite{vysotska2015efficient}   & 
      0.65/-/- &
      0.26/-/-\\

         SeqNet ($S_1$) + GRH:DH~\cite{vysotska2017relocalization}   & 
      0.01/-/- &
      0.09/-/-\\
      
    SeqNet ($S_1$) + GRH:LSH~\cite{vysotska2017relocalization}   & 
      0.34/-/- &
      0.18/-/-\\

\midrule  
\textbf{Hierarchical:} & & \\
        HVPR (NetVLAD to $S_1$) & 
         0.71/0.80/0.85&
        0.25/0.33/0.41\\
        
        HVPR (Delta to $S_1$) & 
         0.65/0.72/0.74&
        0.26/0.39/0.50\\

        HVPR ($S_5$ to $S_1$) & 
        \textbf{0.72}/\textbf{0.82}/\textbf{0.88} &
        \textbf{0.29}/\textbf{0.40}/\textbf{0.55}\\
        
        \bottomrule
    \end{tabular}
    \label{tab:perfComp}
\end{table}

\begin{table}
    \caption{Computation Time (in ms): Mean $\pm$ Std. Dev.}
    \centering
\begin{tabular}{cccc}
    \toprule
    $S_1$ & $S_5$ & SeqMatch: All & SeqMatch: Top 20 \\
    \midrule
     22.1 $\pm$ 13.0 &
     24.2 $\pm$ 11.9 & 
     1449.9 $\pm$ 120.5 & 
     1.8 $\pm$ 0.2\\
    \bottomrule
\end{tabular}
    \label{tab:computeTime}
\end{table}

\begin{table}
\caption{Reverse Traverse - Oxford-FD (Day vs Night): Recall@K}
    \centering
    \begin{tabular}{l cc}
        \toprule
        {} & \multicolumn{2}{c}{\textbf{Recall @ 1/5/20}} \\
        \cmidrule{2-3}
        \textbf{Method}  & \textit{Vanilla SeqMatch} & \textit{Reverse SeqMatch} \\
        \midrule
        
        Raw Single Descriptor  & 
 0.47/0.70/0.85 &
 -\\

        SeqNet ($S_1$)  & 
        0.47/0.69/0.86 &
        -\\

        Raw + Smoothing   & 
        0.59/0.72/0.85 &
        -\\

        Raw + Delta  & 
        0.17/0.32/0.52&
        -\\

        SeqNet ($S_5$)  &  
        0.58/0.73/0.88&
        -\\

        Raw + SeqMatch   & 
        0.49/0.69/0.85 &
        0.67/0.78/0.90\\

         SeqNet ($S_1$) + SeqMatch   & 
       0.58/0.77/\textbf{0.90}	&
       \textbf{0.73}/\textbf{0.85}/\textbf{0.94}	\\

        HVPR ($S_5$ to $S_1$) & 
        \textbf{0.59}/\textbf{0.78}/0.87 &
        0.71/0.81/0.87	\\

        \bottomrule
    \end{tabular}
    \label{tab:oxReverseDb}
\end{table}

\subsection{Reverse Traverse}
Our proposed method SeqNet uses temporal convolutions and sequence averaging which enables learning temporal relations beyond merely memorizing the sequential order of observed information. Unlike concatenation~\cite{arroyo2015towards,facil2019condition} and recurrence~\cite{facil2019condition,neubert2019neurologically} based place descriptions, $S_{L_d}$ does not impose \textit{strict} sequential order. Table~\ref{tab:oxReverseDb} presents results on the Oxford-FD test set with the reference database processed in the reverse order while the query traverse remains as such. We consider two settings here: Vanilla SeqMatch as per Equation~\ref{eq:seqMatch} and Reverse SeqMatch where sequence score aggregation is done in a reverse order to neutralize the effect of reversed database. The latter is considered in particular to observe performance variations in HVPR due to reversed sequential descriptor alone where the effect of reversing in sequence score aggregation is nullified. It can be observed that the performance of the learnt sequential descriptor ($S_5$) does not degrade much (from 0.62 to 0.58) when compared to its counterpart in Table~\ref{tab:perfComp}, despite not having explicitly trained on reverse sequences. On the other hand, sequence score aggregation (Raw + SeqMatch, $S_1$ + SeqMatch and HVPR) suffers significant performance loss due to the reversed order of database images. However, with Reverse SeqMatch, it can be observed that the proposed HVPR pipeline achieves similar performance as compared to its counterpart in Table~\ref{tab:perfComp}.

\subsection{Computational Gain}
Our proposed SeqNet based HVPR pipeline achieves superior recall performance while also reducing the additional compute typically required for sequence score aggregation when considering the whole reference database. The computational complexity in this case is typically $\mathcal{O}(n)$, being a linear multi-variate function $f(N,L,D)$, where $N$ is the number of reference database places (either $N_{db}$ or $K$), $L$ is the sequence length and $D$ is the descriptor dimension. Since $D$ remains constant for a fixed descriptor type, the compute time only varies with the choice of $N$ and $L$. For whole-database sequence matching, compute time is proportional to $N_{db}L_m$, whereas for HVPR, it is proportional to $N_{db}+KL_m$.
For the Nordland test set with $N_{db}=3000$, $K=20$ and $L_m=5$, number of descriptor comparisons for HVPR are only $4.84\%$ of the whole-database sequence matching ($3100 \ll 15000$). 
Note that the use of ANN (Approximate Nearest Neighbor) search techniques for searching single image descriptors can reduce the overall complexity but they mostly operate independently on single image descriptors~\cite{vysotska2017relocalization, harwood2016fanng, garg2020fast} without access to additional sequential information and are thus prone to perceptual aliasing. This is not the case with more robust sequential descriptors which can also benefit from ANN search.

Table~\ref{tab:computeTime} shows GPU (Nvidia GeForce GTX 1080 Ti) computation time for the Boston dataset ($14000$ database images) for different components of the pipeline: descriptor extraction for $S_1$ and $S_5$ (excluding NetVLAD extraction) and sequence score aggregation for the whole database (SeqMatch: All) and HVPR (SeqMatch: Top 20). For this analysis, code optimization based on loop unrolling or array broadcasting was not considered for SeqMatch due to limited GPU RAM.

\section{CONCLUSION}
\label{DISCUSSION}
Learning sequential descriptors using temporal convolutions provides a powerful means by which to generate an initial list of high quality place match hypotheses, with some additional benefits of possessing limited order invariance. When these hypotheses are used for selective sequence matching or filtering of learnt single image descriptors, the result is state-of-the-art performance with reduced overall compute. Future work will address a number of interesting issues identified by this work, in particular the wide variation in how well a sequential description or matching process improves performance of a single image descriptor. Further research could investigate how to predictively quantify the likely responsiveness of a single image descriptor to sequential filtering, and in turn develop a learning framework for optimizing responsiveness driven by the functional requirements of the application like desired matching latency.







\bibliographystyle{IEEEtran}
\bibliography{reflist,moreRefs,refSeqNet}

\end{document}


\maketitle
\thispagestyle{empty}
\pagestyle{empty}

Here, we present additional necessary details, results, and visualizations which could not fit into the main paper but are valuable for any re-implementation and deeper insights.

\section{Experimental Setup: Additional Information}
\subsection{Data Splits}
For each of the datasets, train, validation and test splits were defined without any geographical overlap to observe generalization. The image counts for reference and query databases for train, val and test are presented in Table~\ref{tab:splitInfo}. As the Nordland dataset is unique in its environment-type and spans across multiple cities in its long-route journey, we use Summer-Winter for training and validation, and use Spring-Fall for testing. For the Urban City (Day vs Night) datasets, having access to the day-night data explicitly from different cities, we use the same validation set (from Brisbane) for both the cities (Brisbane City Loop and Oxford Robotcar). For MSLS, no splits were performed within any city and models trained on just one city (Melbourne) were used for testing across 4 other cities, and one city (Austin) was used for validation. However, to limit the training time, we limited the training set to a maximum of 5,000 images.

\begin{table}[b]
    \caption{Data Splits: Reference / Query Database Size}
    \centering
    \tabcolsep=0.11cm
    \begin{tabular}{lllll}
    \toprule
    \textbf{Split} & Oxford & Brisbane & Nordland & MSLS  \\
    \midrule
       Train & 2981 / 2931 & 12625 / 13932 & 15000 / 15000 & 4973 / 4474 \\ 
       Val & -  & 500 / 488 & 3000 / 3000 & 4927 / 1732 \\ 
        Test & 1574 / 1576 & 2858 / 2770 & 3000 / 3000 & 33000 / 18000\\ 
    \bottomrule
    \end{tabular}
    \label{tab:splitInfo}
\end{table}

\subsection{Training Parameters}
We used a margin value of $\alpha=0.3$ for computing the loss which was then minimized using SGD optimizer, with weight decay rate of $0.001$ and momentum $0.9$. The initial learning rate was set to $0.0001$ which was reduced by a factor of $0.5$ every $50$ epochs. For the Oxford Robotcar dataset, we ran training for only 60 epochs and for other larger datasets, Brisbane City Loop, Nordland and MSLS, training was done for $200$ epochs (this is due to the increased number of negatives in proportion to the size of the database as we only consider $10$ negatives for each query~\cite{arandjelovic2016netvlad}). For generating positives/negatives for the triplet loss, we used a maximum/minimum distance of $5/20$ meters for the city datasets and $10/40$ frames for the Nordland dataset. For city datasets, we used $L_d$ as $5$ and $w$ as $3$ for training. For the Nordland dataset, these values were set to $10$ and $5$ respectively. During testing, we used a sequence length of $5$ for all datasets and all methods.

\subsection{Fixed Image Resolution}
As also mentioned in the main paper, we use a fixed image resolution of $640\times 320$ to compute single image NetVLAD descriptors and subsequently extract sequential descriptors from SeqNet. While one could use any image resolution for this purpose, we observed that using different image resolutions between the train and test set led to performance deterioration.

\section{Additional Results and Visualizations}
\subsection{Visualizing Descriptor Space}
Figure~\ref{fig:tsne} shows TSNE~\cite{maaten2008visualizing} visualization of different single image and sequential descriptors for the Nordland Summer test set using first $1000$ images. It can be observed that the sequential descriptors (Smoothing, Delta and SeqNet ($S_5$)) show temporal coherence in the descriptor space. However, neither a too-close packing (as in Delta) nor a too-long temporal coherency (as in Smoothing) is as useful when considering VPR using shorter sequences. It seems that a more desired behavior is to have temporally-coherent descriptors in shorter temporal windows and yet be reasonably far apart from other such clusters, as observed in the visualization of SeqNet ($S_5$) descriptors (last column).

\newcommand{\scaleTwo}{0.175\textwidth}

\begin{figure*}
    \centering
    \begin{tabular}{ccccc}
    \includegraphics[width=\scaleTwo]{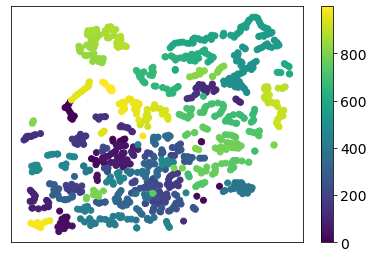} &
    \includegraphics[width=\scaleTwo]{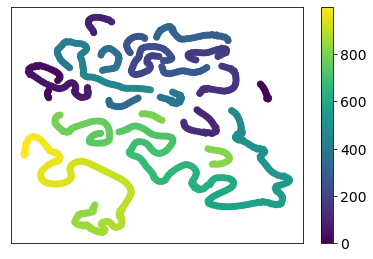} &
    \includegraphics[width=\scaleTwo]{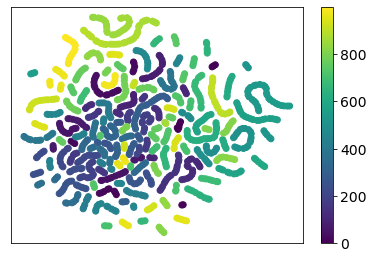} &
    \includegraphics[width=\scaleTwo]{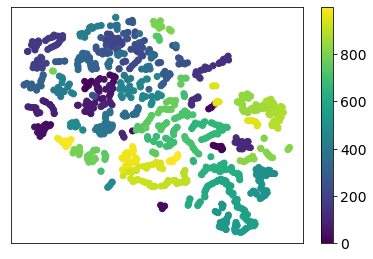} &
    \includegraphics[width=\scaleTwo]{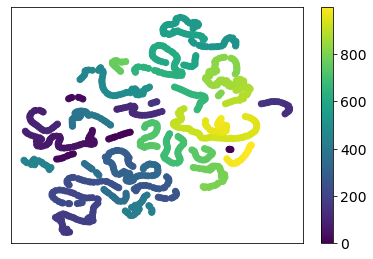} \\
    Raw Single Descriptor & Raw + Smoothing & Raw + Delta & SeqNet ($S_1$) & SeqNet ($S_5$) \\
    \end{tabular}
    \caption{TSNE visualization of different descriptor types for Nordland Summer test set using first 1000 images.}
    \label{fig:tsne}
\end{figure*}

\subsection{Longer Sequences}
In our baseline comparisons in the main paper, we only considered a sequence length of 5 for our proposed methods as well as for the baselines. This was motivated by the application of VPR in global relocalization with reduced latency as opposed to long-term continuous integration of sequential information which necessitates exact route repetition for longer distances. In Table~\ref{tab:longerSeq}, using Oxford-FD (Day-Night) test set and Brisbane (Day-Night) trained model, we compare HVPR with GISM~\cite{vysotska2015efficient} for longer sequence lengths ($10$ and $20$), considering the latter's inherent design to process sequential information in a continuous long-term manner. For HVPR, we keep the length of sequential descriptor $S_{L_d}$ as $5$ and leverage additional sequential information in the second stage of sequence score aggregation (SeqMatch), that is, vary $L_m$ to $10$ and $20$. It can be observed in Table~\ref{tab:longerSeq} that both GISM and HVPR benefit from an increasing sequence length while HVPR outperforms GISM for all sequence lengths.

\begin{table}[t]
\caption{Longer Sequences - Oxford-FD (Day vs Night): Recall@K}

    \centering
    \begin{tabular}{l lll}
        \toprule
        {} & \multicolumn{3}{c}{\textbf{Recall @ 1/5/20}} \\
        \cmidrule(lr{0.75em}){2-4}
        \textbf{Method}  & $L_m=5$ & $L_m=10$ & $L_m=20$ \\
        \midrule
        
        GISM~\cite{vysotska2015efficient}   &
        0.65/-/- &
        0.68/-/- &
        0.73/-/- \\
        
        HVPR & 
        \textbf{0.72}/\textbf{0.82}/\textbf{0.88} &
        \textbf{0.80}/\textbf{0.85}/\textbf{0.88} &
        \textbf{0.85}/\textbf{0.87}/\textbf{0.88} \\

        \bottomrule
    \end{tabular}
    \label{tab:longerSeq}
\end{table}

\subsection{Different Underlying Single Image Descriptor}
We used NetVLAD~\cite{arandjelovic2016netvlad} as the underlying single image descriptor for all our proposed methods. However, the proposed sequential descriptors and the HVPR pipeline is not tied to this particular choice. Table~\ref{tab:nordGMP} shows results on Nordland's Spring-Fall test set with a single image descriptor of another type: Global Max Pool (GMP) descriptor (similar to MAC proposed in~\cite{tolias2015particular}), obtained from the final convolutional layer of ResNet50~\cite{he2016deep} trained on object recognition task. It can be observed that the overall performance trends for GMP are similar to that achieved by NetVLAD. It can also be observed that the absolute performance achieved through the proposed HVPR pipeline is similar for both the underlying single image descriptors even though GMP is half the size of NetVLAD ($2048$ vs $4096$). Furthermore, it can be observed that HVPR based on GMP surpasses NetVLAD even though GMP's baseline performance (Raw Single Descriptor and Raw + SeqMatch) is much worse than NetVLAD. This observation hints that a different set of visual attributes might be more relevant when learning from temporal information than the ones relevant for learning single image descriptors. 

\begin{table}[h]
\caption{Different Single Image Descriptor - Nordland (Spring vs Fall): Recall@K}
    \centering
    \begin{tabular}{l cc}
        \toprule
        {} & 
        \multicolumn{2}{c}{\textbf{Recall @ 1/5/20}} \\
        \cmidrule{2-3}
        \textbf{Method} & 
        \textit{NetVLAD} & \textit{GMP} \\
        \midrule
        
        Raw Single Descriptor & 
 0.38/0.54/0.68 &
 0.36/0.54/0.68	\\

        SeqNet ($S_1$) & 
        0.48/0.69/0.82 &
0.49/0.71/0.83 \\

        Raw + Smoothing  & 
        0.44/0.59/0.72  &
 0.37/0.52/0.65	\\   

        Raw + Delta  & 
        0.56/0.70/0.80	&
0.49/0.62/0.76	\\

        SeqNet ($S_5$) & 
        \textbf{0.79}/\textbf{0.90}/\textbf{0.94}&
        0.76/0.88/\textbf{0.94}	\\

        Raw + SeqMatch  & 
        0.61/0.71/0.78	&
0.57/0.66/0.76	\\    
        
        SeqNet ($S_1$) + SeqMatch  & 
       0.78/0.87/0.92	&
        0.79/0.86/0.91	\\

        HVPR ($S_5$ to $S_1$)  & 
        \textbf{0.79}/0.89/\textbf{0.94}	&
        \textbf{0.82}/\textbf{0.90}/\textbf{0.94} \\

        \bottomrule
    \end{tabular}
    \label{tab:nordGMP}
\end{table}

\bibliographystyle{IEEEtran}
\bibliography{reflist,moreRefs,refSeqNet}